\documentclass[11pt,letterpaper]{article}
\usepackage{emnlp2017}
\usepackage{times}
\usepackage{latexsym}
\usepackage{graphicx}
\usepackage[utf8]{inputenc}
\usepackage{amssymb}

\emnlpfinalcopy

\def\emnlppaperid{***}

\title{Neural Machine Translation with Extended Context}

 \author{J{\"o}rg Tiedemann \and Yves Scherrer\\ Department of Modern
     Languages \\ University of Helsinki}

\date{}

\begin{document}

\maketitle


\begin{abstract}
We investigate the use of extended context in attention-based neural machine translation. We base our experiments on translated movie subtitles and discuss the effect of increasing the segments beyond single translation units. We study the use of extended source language context as well as bilingual context extensions. The models learn to distinguish between information from different segments and are surprisingly robust with respect to translation quality. In this pilot study, we observe interesting cross-sentential attention patterns that improve textual coherence in translation at least in some selected cases.
\end{abstract}

\section{Introduction}

Typical models of machine translation handle sentences in isolation and discard any information beyond sentence boundaries. Efforts in making statistical MT aware of discourse-level phenomena appeared to be difficult \cite{Hardmeier:2012b,Carpuat:2012,Hardmeier:2013}. Various studies have been published that consider textual coherence, document-wide translation consistency, the proper handling of referential elements such as pronominal anaphora and other discourse-level phenomena \cite{Guillou:2012,Russo:2012,Voigt:2012,Xiong:2013a,Ben:2013,Xiong:2013b,Xiong:2013,LOAICIGA14.205.L14-1205}. The typical approach in the literature focuses on the development of task-specific components that are often tested as standalone modules that need to be integrated with MT decoders \cite{Hardmeier:2013a}. Modest improvements could, for example, be shown for the translation of pronouns \cite{Le-Nagard:2010,Hardmeier:2010,Hardmeier:2014a}  and the generation of appropriate discourse connectives \cite{Meyer:2012a}. Textual coherence is also often tackled in terms of translation consistency for domain-specific terminology based on the one-translation-per-discourse principle \cite{Carpuat:2009,Tiedemann:2010a,Ma:2011,Ture:2012}.

Overall, none of the ideas lead to significant improvements of translation quality. Besides, the development of task- and problem-specific models that work independently from the general translation task is not very satisfactory. However, the recent success of neural machine translation opens new possibilities for tackling discourse-related phenomena in a more generic way. In this paper, we present a pilot study that looks at simple ideas for extending the context in the framework of standard attention-based encoder-decoder models. The purpose of the paper is to identify the capabilities of NMT to discover cross-sentential dependencies without explicit annotation or guidance. 
In contrast to related work that modifies the neural MT model by an additional context encoder an a separate attention mechanism \cite{2017arXiv170405135J}, we keep the standard setup and just modify the input and output segments. 
We run a series of experiments with different context windows and discuss the effect of additional information on translation and attention.

\section{Attention-Based NMT}

Encoder-decoder models with attention have been proposed by \citet{Bahdanau2014nmt} and have become the de-facto standard in neural machine translation. The model is based on recurrent neural network layers that encode a given sentence in the source language into a distributed vector representation that will be decoded into the target language by another recurrent network. The attention model makes 
use of the entire encoding sequence, and the attention weights specify the proportions with which information from different positions is combined. This is a very powerful mechanism that makes it possible to handle arbitrarily long sequences without limiting the capacity of the internal representation. Previous work has shown that NMT models can successfully learn attention distributions that explain intuitively plausible connections between source and target language. This framework is very well suited for the study we conduct in this paper as we emphasise the capabilities of NMT to pick up contextual dependencies from wider context across sentence boundaries.

In our work, we rely on the freely available Helsinki NMT system (HNMT) \citep{hnmt2017}\footnote{\url{https://github.com/robertostling/hnmt}} that implements a hybrid bidirectional encoder with character-level backoff \citep{Luong2016achievingopen} using recurrent LSTM units \citep{Hochreiter1997lstm}. The system also features layer normalisation \citep{Ba2016layernormalization}, variational dropout \citep{Gal2016theoretically}, coverage penalties \citep{Wu2016gnmt}, beam search decoding and straightforward model ensembling. The backbone is Theano, which enables efficient GPU-based training and decoding with mini-batches.

\section{Data Sets}

In our experiments, we focus on the translation of movie subtitles and in particular on the translation from German to English. The choice of languages is rather arbitrary and mainly due to better comprehension for our qualitative inspections. There are relevant discourse phenomena that need to be considered for English and German, for example, referential pronouns with grammatical agreement requirements. The choice of movie subtitles has several reasons: First of all, large quantities of training data are available, a necessary prerequisite for neural MT. Secondly, subtitles expose significant discourse relations and cross-sentential dependencies. Referential elements are common, as subtitles usually represent coherent stories with narrative structures with dialogues and natural interactions. Proper translation in this context typically requires more than just the text but also information from the plot and the audiovisual context. However, as those types of information are not available, we hope that extended context at least helps to incorporate more knowledge about the situation and in consequence leads to better translations, also stylistically. The final advantage of subtitles is the size of the translation units. Sentences (and sentence fragments) are typically much shorter compared to other genres such as newspaper texts or other 
edited written material. Utterances are even shortened substantially for space limitations. This property supports our experiments in which we want to include context beyond sentence boundaries. Similar to statistical MT, neural MT also struggles most with long sequences and, therefore, it is important to keep the segments short. On average there are about 8 tokens per language in each aligned translation unit (which may cover one or more sentences or sentence fragments).

In particular, we use the publicly available OpenSubtitles2016 corpus \citep{LisonTiedemann:2016:LREC} for German and English\footnote{\url{http://opus.lingfil.uu.se/OpenSubtitles2016.php}} and reserve 400 randomly selected movies for development and testing purposes. In total, there are 16,910 movies and TV series in the collection. We tokenized and truecased the data sets using standard tools from the Moses toolbox \citep{Koehn-Hoang-etal:2007}. The final corpus comprises 13.9 million translation units with about 107 million tokens in German and 115 million tokens in English. The training data includes 13.5 million training instances and we selected the 5,000 first translation units of the test set for automatic evaluation. Note that we trust the alignment and do not correct any possible alignment errors in the data.


\begin{figure*}[t]
\centering
{\footnotesize
\begin{tabular}{rl}
\multicolumn{1}{c}{{\em SOURCE}} & \multicolumn{1}{c}{{\em TARGET}}\\
cc\_sieh cc\_, cc\_Bob cc\_! {\bf -Wo sind sie ?} & - Where are they ?\\
cc\_-Wo cc\_sind cc\_sie cc\_? {\bf siehst du sie ?} & do you see them ?\\
cc\_siehst cc\_du cc\_sie cc\_? {\bf -Ja .} & - Yes .\\
\end{tabular}
}
\caption{Example of data with extended source language context.}
\label{fig:srccontext}
\end{figure*}

\begin{figure*}[t]
\centering
{\footnotesize
\begin{tabular}{rl}
\multicolumn{1}{c}{{\em SOURCE}} & \multicolumn{1}{c}{{\em TARGET}}\\
sieh , Bob ! \_BREAK\_ -Wo sind sie ? & look , Bob ! \_BREAK\_ - Where are they ?\\
-Wo sind sie ? \_BREAK\_ siehst du sie ?& - Where are they ? \_BREAK\_ do you see them ?\\
siehst du sie ? \_BREAK\_ -Ja . & do you see them ? \_BREAK\_ - Yes .\\
\end{tabular}
}
\caption{Example of data with extended translation units.}
\label{fig:context}
\end{figure*}

\section{Extended Context Models}

We propose to simply extend the context when training models (and translating data). This does not lead to any changes in the model itself, and we let the training procedures discover what kind of information is needed for the translation. 
We evaluate two models that extend context in different ways:

\begin{description}
\item[Extended source: ] Include context from the previous sentences in the source language to improve the encoder part of the network.
\item[Extended translation units: ] Increase the segments to be translated. Larger segments in the source language have to be translated into corresponding units in the target language.
\end{description}

\paragraph{Model 2+1 (extended source):} In order to keep the segments as short as possible, we will limit ourselves to one contextual unit. Hence, in the first setup, we add the source language sentence(s) from the previous translation unit to the sentence to be translated and mark all tokens (BPE segments in our case) with a special prefix ({\em cc\_}) to indicate that they come from contextual information. We also test a second model without prefix-marked context words but additional sentence-break tokens between the source language units (similar to model 2+2 below). In that case, we do not make a difference between contextual words and sentence-internal words, which makes it possible to treat intra-sentential anaphora in the same way as cross-sentential ones. 
We run through the training data with a sliding window, adding the contextual history to each sentence in the corpus. Note that we have to make sure that each movie starts without context. Figure~\ref{fig:srccontext} shows a few examples from our test set with the prefix-markup described above.

The task now consists in learning the influence of specific context word sequences on the translation of the focus sentence. An example is the ambiguous pronoun ``sie'' that could be a feminine singular or a plural third person pronoun. The use of grammatical gender in German also makes it possible to refer to an inanimate antecedent. Discourse-level information is needed to make correct decisions. The question is whether our model can actually pick this up and whether attention patterns can show the relevant connections.

\paragraph{Model 2+2 (extended translation units):} In the second setup, we simply add the previous translation unit to extend context in both source and target during training. With this model, the decoder also has to generate more content but is probably less likely to confuse information from different positions as it simply translates larger units. Another advantage is that target-language-specific dependencies like grammatical agreement between referential expressions may be captured if they cannot be determined by the source language alone. As above, we run through the training data with a sliding window and create extended training examples, marking the boundaries between the segments with a special token {\em \_BREAK\_}. Figure~\ref{fig:context} shows the example from the test data.

\vspace{1em}
The NMT models that we train rely on subword-units. We apply standard byte-pair encoding (BPE) \citep{Sennrich2016subword} for splitting words into segments. For the extended source context models, we set a vocabulary size of 30,000 when training BPE codes and apply a vocabulary size of 60,000 when training the models (context words double the vocabulary because of their {\em cc\_} prefix). For the 2+2 model, we train BPE codes from both languages together (with a size of 60,000) and we set a vocabulary threshold of 50 when applying BPE to the data.

\section{Experiments and Results}

We train attention-based models using the Helsinki NMT system with 
similar parameters but different training data to see the effect of contextual information. Our baseline system involves a standard setup where the training examples come from the aligned parallel subtitle corpus (1 source translation unit and 1 target translation unit). This will be the reference in our evaluations and discussions. In all cases, we translate the test set of 5,000 sentences with an ensemble model consisting of the final four savepoint models after running roughly the same number of training iterations with similar amounts of training instances seen by the model. Savepoint averaging slightly alleviates the problem that each model will differ due to the stochastic nature of the training procedures, making a direct comparison of the outcomes difficult especially if the observed differences are small. 

Automatic evaluation metrics are problematic, in particular for assessing discourse-related phenomena. However, it is important to verify that the context-models are on-par with the baseline. Table~\ref{tab:BLEU} shows the BLEU scores and also the alternative character-level chrF3 measure for all systems (2+1 in its two variants with and without prefix markup). The 2+2 model is evaluated on the last segment in the generated output and ignores all other parts before.

\begin{table}[ht]
\centering
{\small
\begin{tabular}{|l|cccc|}
\hline
in \%            & BLEU &   chrF3 & (precision) & (recall)\\
\hline
baseline       & 27.1 & 42.9 & {\bf 54.7} & 41.9\\ 
2+1 (prefix) & 26.5 & 42.7 & 51.2 & 41.9 \\
2+1 (break) & {\bf 27.5} & {\bf 43.3} & 52.8 & {\bf 42.5} \\
2+2             & 26.5 & {\bf 43.3} & 54.4 & 42.3\\
\hline
\end{tabular}
}
\caption{Automatic evaluation: BLEU and chrF3 (including precision and recall).}
\label{tab:BLEU}
\end{table}

The table shows that all models are quite similar to each other, with a slightly higher BLEU score for the 2+1 system with sentence breaks. The chrF3 score is also slightly higher for both, the 2+1 and 2+2 systems with sentence breaks, due to a higher recall. The differences are small but the results already show that the system is capable of handling larger units without harming the performance and additional improvements are possible. Let us know look at some details to study the effects of contextual information on translation output.


\subsection{2+1: Extended Source Language Context}

The most difficult part for the model in the 2+1 setup is to learn to ignore most of the contextual information when generating the target language output. In other words, the attention model needs to learn to focus on words and word sequences that are relevant to the translation process. It is interesting to see that the system is actually able to do that and produce adequate translations even though a lot of extra information is given in the source. There is certainly some confusion in the beginning of the training process but the model figures out surprisingly quickly what kind of information to consider and what information to discard.

It is interesting to see, of course, how much of the contextual information is still used and where. For this, we looked at the distribution of attention in the whole data set, for individual sentences and for individual target words. The total proportion of attention that goes to the contextual history is about 7.1\%. This is small -- as expected -- but certainly not negligible. When sorting by contextual attention, some sentences actually show quite high proportions of attention going to the previous context. They mainly refer to translations that include information from the previous history or rather creative translations that are less faithful to the original source. An example is given below (context in parentheses):

\medskip

\noindent
{\footnotesize
\begin{tabular}{|rp{6cm}|}
\hline
input &  (Danke , Mr. Vadas .) {\bf Mr. Kralik , kommen Sie bitte ins Büro . ich möchte Sie sprechen .}\\
transl. & Mr. Kralik , please come to the office , I want to talk to you .\\
\hline
input & (Mr. Kralik , kommen Sie bitte ins Büro . ich möchte Sie sprechen .) {\bf ja .}\\
transl. & Yes , I want to speak to you .\\
\hline
\end{tabular}
}

\medskip

\begin{figure}[ht]
\includegraphics[width=\columnwidth]{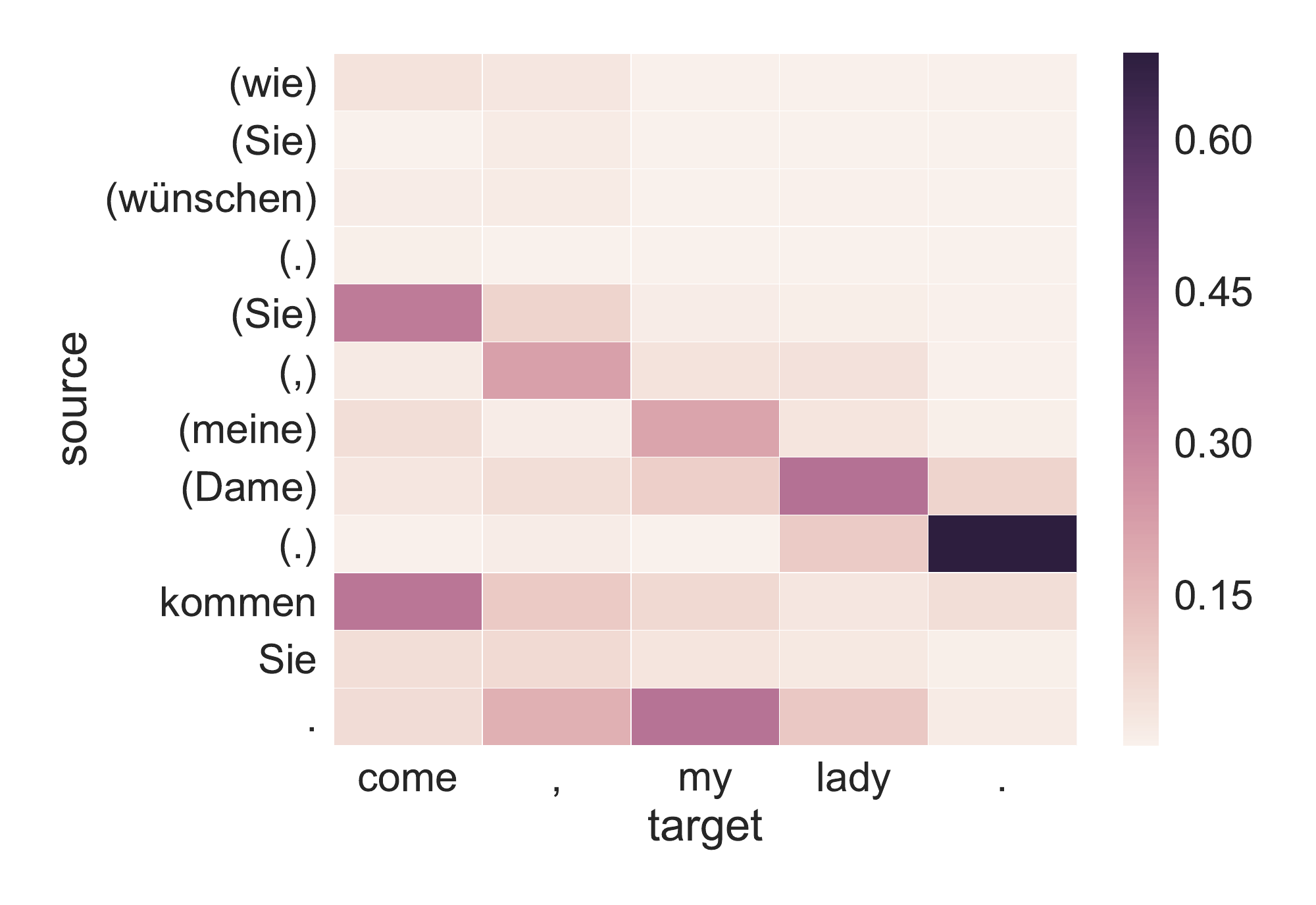}
\caption{Attention with extended source context. Words from the contextual history are in parentheses.}
\label{img:2+1-41}
\end{figure}

The second sentence to be translated ({\em ``ja .''}) is filled with a repetition from the contextual history. The part {\em ``I want to speak to you''} is indeed mostly linked to the German {\em ``ich möchte Sie sprechen''} from the history. Such repetitions may feel quite natural (for example, if the speaker is the same and would like to stress the previous request) and one is tempted to say that the model picks this possibility up from the data where such examples occur. 
However, such cases seem to occur especially in connection with multiple sentences in the source context. The following translation illustrates another interesting case with two context sentences. Figure~\ref{img:2+1-41} shows the attention pattern in which the model replaced the referential {\em ``Sie''} from the source sentence by {\em ``my lady''} from the previous context.


Similarly, the following example shows again how information from the context is merged with the current sentence to be translated:

\medskip

\noindent
{\footnotesize
\begin{tabular}{|rp{6cm}|}
\hline
input &  (Pirovitch .) \newline {\bf - Hm ? -  Wollen Sie was Nettes hören ?}\\
transl. & - You want to hear something nice ?\\
\hline
input & (- Hm ? -  Wollen Sie was Nettes hören ?) \newline {\bf Was denn ?}\\
transl. & - What do you want to hear ?\\
\hline
\end{tabular}
}

\medskip

\begin{figure}[ht]
\hspace{-7pt}\includegraphics[width=1.1\columnwidth]{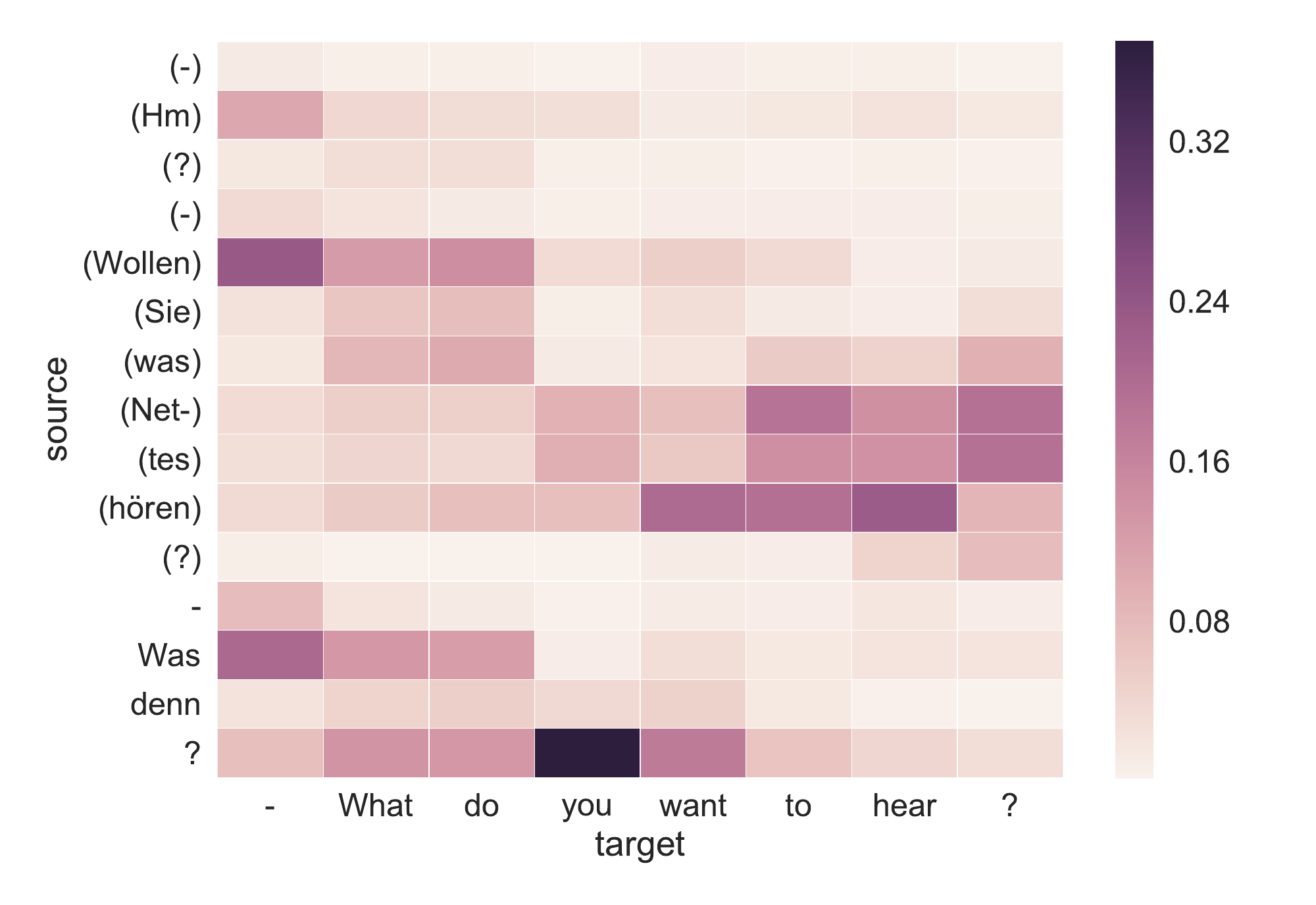}
\caption{Another example for attention with extended source context.}
\label{img:21-3147}
\end{figure}

The attention heatmap in Figure~\ref{img:21-3147} nicely illustrates how the translation picks up from the conversation history. Once again, this kind of mix could be possible if the speaker stays the same but, probably, this is not the case and the translation is altered in such a way that it becomes incorrect in this context. These observations suggest that additional information such as speaker identities or dialog turns will be necessary to handle such cases correctly.

The examples above constitute rather anecdotal evidence and systematic patterns are difficult to extract. We leave it to future work to study various cases in more detail and to inspect certain properties in connection with specific discourse phenomena. In this paper, we inspect instead the distribution of attention for individual target words to see what word types depend most on contextual history. For this, we counted the overall attention of each word type in our test set and computed the proportion of external attention on average. The list of the top ten words (after lowercasing) with frequency above four is given in Table~\ref{tab:external-attn1}. Those words receive considerably larger external attention (17-26\%) than the average (4.9\%).

\begin{table}[ht]
\centering
{\small
\begin{tabular}{|l|@{\hskip5pt}r@{\hskip5pt}c@{\hskip5pt}c@{\hskip5pt}r@{\hskip5pt}c|}
\hline
word & freq & external & internal & prop.\% & $\varnothing$ pos.\\
\hline
yeah & 35 & 0.224 & 0.622 & 26.5 & 3.71\\
yes & 182 & 0.212 & 0.601 & 26.1 & 4.22\\
wake & 6 & 0.239 & 0.684 & 25.8 & 6.67\\
anywhere & 6 & 0.223 & 0.655 & 25.4 & 7.67\\
course & 35 & 0.191 & 0.631 & 23.2 & 3.17\\
oh & 61 & 0.199 & 0.712 & 21.9 & 2.08\\
saying & 5 & 0.177 & 0.690 & 20.5 & 5.20\\
tired & 9 & 0.174 & 0.774 & 18.3 & 5.67\\
latham & 5 & 0.169 & 0.796 & 17.5 & 7.80\\
really & 13 & 0.161 & 0.763 & 17.4 & 2.77\\
\hline 
{\em average} & --- & 0.045 & 0.891 & 4.9 & --- \\
\hline
(36) she  &   98  &    0.124  & 0.837  & 12.9 & 3.70\\
(62) he    & 232  &   0.103  & 0.851  & 10.8 & 4.04\\
(79) it    &533    & 0.089   &0.807  & 10.0 & 4.81\\
(83) they  & 135   &  0.095  & 0.871  & 9.9 & 4.17\\
(97) you   &1349   & 0.084  & 0.828 &  9.2 & 4.28\\
\hline
\end{tabular}
}
\caption{Word types with the highest external attention and the rank of some cross-lingually ambiguous pronouns in the list sorted by the proportion ({\em prop.}) of external attention. {\em $\varnothing$ pos.} gives the average token position of the target word.}
\label{tab:external-attn1}
\end{table}

Unfortunately, there is no straightforward interpretation of the words that receive substantial attention from the extended contextual history, but several response particles such as ``yes'', ``yeah'', ``oh'', which glue together interactive dialogues, are in the list. Furthermore, we can see that the words with significant cross-sentential attention does not consist of sentence-initial words only. The token position varies quite a lot. 
We also list the values of pronouns with significant cross-lingual ambiguity and their rank in the list sorted by the proportion of external attention. The third-person pronouns ``he'', ``she'' and ``it'' put significant attention (over 10\%) on the previous sentence(s).

Some words are simply not easy to link to particular source language words and, therefore, their attention may be spread all over the place. Therefore, we also computed the proportion of external attention at specific positions in the input by considering only the highest internal and the highest external attention for each target word in each sentence. The list of words with the highest external attention according to that measure are listed in Table~\ref{tab:external-attn2}.

\begin{table}[ht]
\centering
{\small
\begin{tabular}{|l|@{\hskip5pt}r@{\hskip5pt}c@{\hskip5pt}c@{\hskip5pt}r@{\hskip5pt}c|}
\hline
word & freq & external & internal & prop.\% & $\varnothing$ pos.\\
\hline
yeah & 35 & 0.135 & 0.242 & 35.9 & 3.71\\
wake & 6 & 0.179 & 0.326 & 35.4 & 6.67\\
yes & 182 & 0.091 & 0.259 & 26.1 & 4.22\\
tired & 9 & 0.113 & 0.364 & 23.7 & 5.67\\
oh & 61 & 0.086 & 0.288 & 23.1 & 2.08\\
anywhere & 6 & 0.094 & 0.326 & 22.4 & 7.67\\
dover & 6 & 0.119 & 0.426 & 21.9 & 5.83\\
course & 35 & 0.069 & 0.271 & 20.3 & 3.17\\
speak & 15 & 0.072 & 0.305 & 19.1 & 4.67\\
sure & 29 & 0.065 & 0.284 & 18.7 & 3.59\\
\hline 
{\em average} & --- & 0.021 & 0.441 & 4.5 & ---\\
\hline
(20) she & 98 & 0.062 & 0.343 & 15.3 & 3.70\\
(50) he & 232 & 0.048 & 0.355 & 11.9 & 4.04\\
(64) it & 533 & 0.040 & 0.332 & 10.8 & 4.81\\
(72) you & 1349 & 0.038 & 0.327 & 10.4 & 4.28\\
(79) they & 135 & 0.040 & 0.356 & 10.1 & 4.17\\
\hline
\end{tabular}
}
\caption{Word types with the highest average of external attention peaks.}
\label{tab:external-attn2}
\end{table}

The list is quite similar to the previous one, but one notes that the pronouns all advance in the rankings, suggesting a more focused attention of these entities. This is an interesting observation, and we will leave further investigations to future work.

\subsection{2+2: Larger Translation Units}

Let us turn now to the second model that works with larger translation units. Here, the neural network produces a translation of the entire extended input. This includes the generation of segment break symbols and attention for the entire sequence. Again, the question arises whether the model learns to look at information outside of the aligned segment. External context is not marked with specific prefixes anymore and token representations are completely shared in the model. Theoretically, the model can now swap, shuffle or merge information that comes from different segments. Random inspection does not yield many such cases, but we do see a number of cases where translations include information from previous parts or where the segment break is placed in a different position than in the reference translation. Often, this is actually due to alignment errors in the reference data, such that the translation system is penalised without reason in our automatic evaluation. Table~\ref{tab:BLEU2} shows scores of the extended context translations and we can now see a slight improvement in BLEU and chrF3. Note that each translation hypothesis and each reference now refers to two segments with break tokens between them removed. Hence, the scores do not match the ones in Table~\ref{tab:BLEU}.

\begin{table}[ht]
\centering
{\small
\begin{tabular}{|l|cccc|}
\hline
in \%            & BLEU &   chrF3 & (precision) & (recall)\\
\hline
baseline*     & 27.25 & 44.14 & 55.61 & 43.15\\
2+2*           & 27.41 & 44.54 & 55.51 & 43.58\\
\hline
\end{tabular}
}
\caption{BLEU and chrF3 on extended context segments (sliding window). Individual segments are simply concatenated in the baseline system where necessary.}
\label{tab:BLEU2}
\end{table}



\begin{figure}[ht]
\hspace{-10pt}\includegraphics[width=1.1\columnwidth]{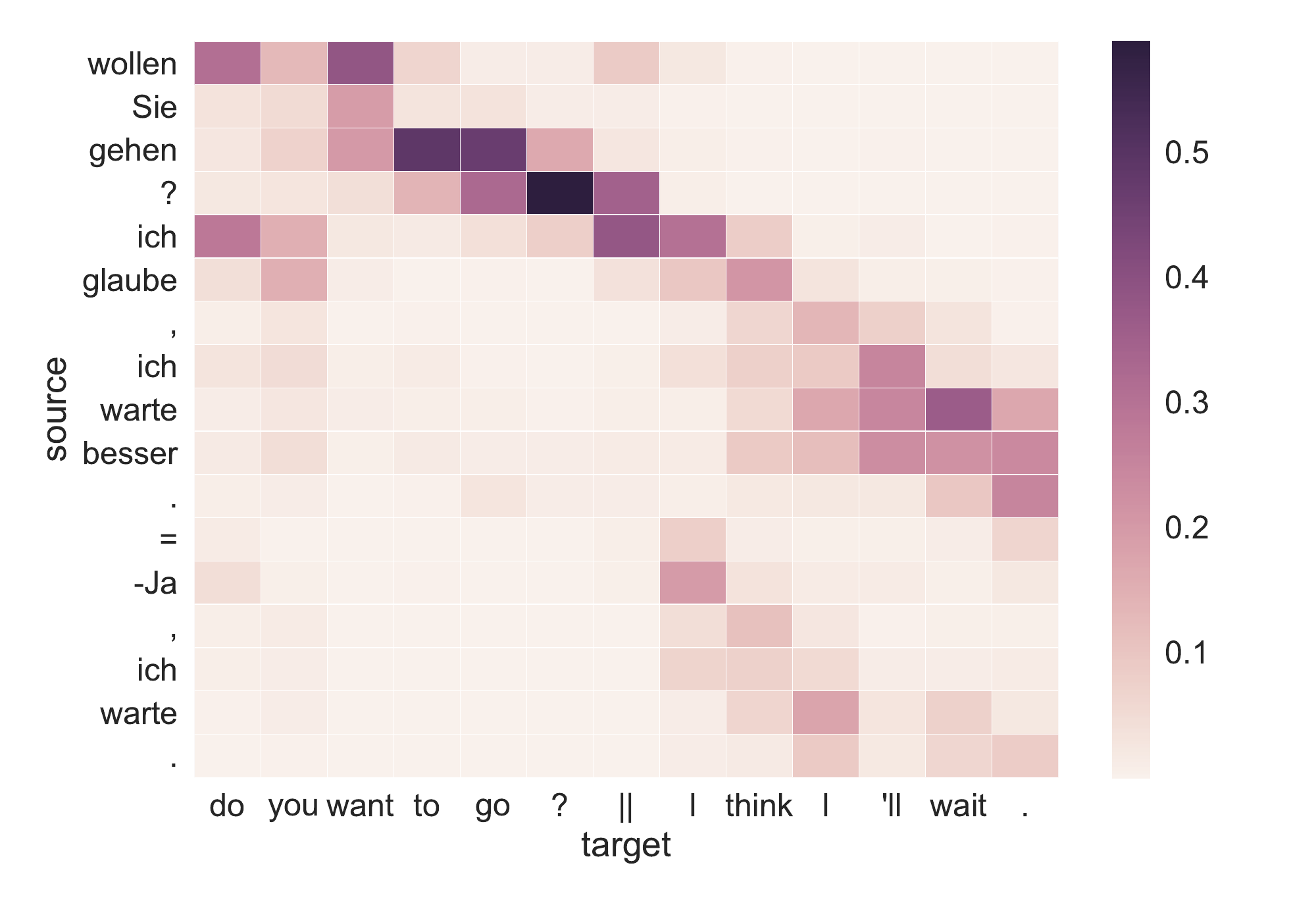}
\caption{Attention with multiple sentences and large cross-segment attention. The double bars refer to segment breaks.}
\label{img:2+2-1324}
\end{figure}

Figure~\ref{img:2+2-1324} illustrates an example with a large proportion of cross-segmental attention. In this case, the model summarises part of segment one with segment two into one translation, and the attention goes mainly to segment one.

This looks quite acceptable from the point of view of coherence. Looking at the reference used for automatic evaluation, we can actually see a misalignment in the data where {\em ``do you want to go ?''} should have been aligned to {\em ``wollen Sie gehen ?''}:

\medskip

\noindent
{\footnotesize
\begin{tabular}{|l|}
\hline
I don 't care what you 've started . do you want to go ?\\
mir ist egal , was sie angefangen haben . \\
\hline
no , I think I 'd better wait .\\
wollen Sie gehen ? ich glaube , ich warte besser . \\
\hline
- Yes , I 'll wait .\\
-Ja , ich warte . \\
\hline
\end{tabular}
}

\medskip

%

It is also interesting to see that the generation of the segment break symbol uses information from segment-initial tokens and punctuations such as question marks. This also follows the intuitions about the decision whether a segment is complete or not.

We also computed word-type-specific attention again. However, the list of words that put significant focus on other segments looks quite different from the previous model. The top-ten list is shown in Table~\ref{tab:external-attn1-22}.

\begin{table}[ht]
\centering
{\small
\begin{tabular}{|l|@{\hskip5pt}r@{\hskip5pt}c@{\hskip5pt}c@{\hskip5pt}r@{\hskip5pt}c|}
\hline
word & freq & external & internal & prop.\% & $\varnothing$ pos.\\
\hline
exactly & 5 & 0.190 & 0.644 & 22.8 & 2.20\\
shelf & 5 & 0.202 & 0.692 & 22.6 & 8.40\\
upstairs & 5 & 0.186 & 0.757 & 19.7 & 7.60\\
unbelievable & 7 & 0.151 & 0.641 & 19.1 & 2.86\\
yeah & 91 & 0.144 & 0.667 & 17.8 & 1.95\\
hardly & 5 & 0.155 & 0.740 & 17.4 & 2.20\\
cares & 5 & 0.144 & 0.755 & 16.0 & 2.60\\
horns & 8 & 0.134 & 0.713 & 15.8 & 5.25\\
fossils & 7 & 0.137 & 0.744 & 15.5 & 3.57\\
-what & 10 & 0.121 & 0.660 & 15.5 & 1.00\\
\hline 
{\em average} & --- & 0.028 & 0.880 & 3.1 & ---\\
\hline
\end{tabular}
}
\caption{Word types with the highest cross-segmental attention (excluding attention on sentence break symbols)).}
\label{tab:external-attn1-22}
\end{table}

We also computed the average attention peak and the proportion of such attention to other segments. The words with highest values are shown in Table~\ref{tab:external-attn2-22}. Again, we can see response particles but also some additional adverbials that can have connective functions. Pronouns appear quite low in the ranked list and, therefore, we leave them out in the presentation here.

\begin{table}[ht]
\centering
{\small
\begin{tabular}{|l|@{\hskip5pt}r@{\hskip5pt}c@{\hskip5pt}c@{\hskip5pt}r@{\hskip5pt}c|}
\hline
word & freq & external & internal & prop.\% & $\varnothing$ pos.\\
\hline
-the & 5 & 0.436 & 0.541 & 44.6 & 1.00\\
-what & 10 & 0.358 & 0.519 & 40.9 & 1.00\\
exactly & 5 & 0.171 & 0.266 & 39.2 & 2.20\\
-aye & 12 & 0.345 & 0.550 & 38.5 & 1.00\\
-yes & 7 & 0.281 & 0.472 & 37.3 & 1.00\\
apparently & 7 & 0.308 & 0.536 & 36.5 & 1.00\\
hardly & 5 & 0.178 & 0.321 & 35.7 & 2.20\\
anyway & 9 & 0.241 & 0.443 & 35.2 & 1.00\\
ah & 6 & 0.217 & 0.407 & 34.8 & 1.00\\
ahoy & 6 & 0.304 & 0.590 & 34.0 & 1.00\\
\hline 
{\em average} & --- & 0.043 & 0.440 & 8.9 & ---\\
\hline
\end{tabular}
}
\caption{Word types with the highest average of cross-segmental attention peaks.}
\label{tab:external-attn2-22}
\end{table}

Cross-segmental attention peaks are dominated by tokens with relatively low overall frequency, some of which arise from tokenization errors (e.g. the words starting with a hyphen, typically from sentence-initial positions). Therefore, we propose another type of evaluation, less sensitive to overall frequency: we only count occurrences of target words whose external attention is higher than the internal attention, and normalize them by the total occurrence count of the target word. We discard words which have majoritarily external attention in four or less cases. Results are shown in Table~\ref{tab:external-attn3-22}.

\begin{table}[ht]
\centering
{\small
\begin{tabular}{|l|rrr|}
\hline
word & proportion & freq ext peak & freq \\
\hline
yeah & 0.077 & 7 & 91 \\
oh & 0.069 & 7 & 101 \\
yes & 0.054 & 11 & 204 \\
thank & 0.049 & 7 & 144 \\
no & 0.025 & 8 & 320 \\
- & 0.023 & 44 & 1890 \\
good & 0.018 & 5 & 284 \\
here & 0.017 & 6 & 346 \\
? & 0.016 & 29 & 1812 \\
... & 0.016 & 5 & 316 \\
. & 0.014 & 104 & 7645 \\
what & 0.012 & 6 & 486 \\
you & 0.009 & 23 & 2458 \\
that & 0.008 & 6 & 725 \\
's & 0.008 & 9 & 1102 \\
it & 0.005 & 5 & 914 \\
, & 0.004 & 16 & 3561 \\
i & 0.004 & 10 & 2372 \\
\hline
\end{tabular}
}
\caption{Word types with the highest proportion of cross-segmental attention peaks, with absolute frequencies of cross-segmental attention peak and overall absolute word frequencies.}
\label{tab:external-attn3-22}
\end{table}

In addition to the known response particles and punctuation signs, we also see pronouns and demonstrative particles (such as \textit{here, what, that}) ranked prominently. However, the absolute numbers are small and only permit tentative conclusions. This analysis also allows us to see the direction of cross-segmental attention. Items that tend to occur at the beginning of the sentence show attention towards the previous sentence, whereas items that occur at the end of a sentence (such as punctuation signs, but also the \textit{`s} token) show attention towards the following sentence.

\begin{figure}[ht]
\hspace{-10pt}\includegraphics[width=1.1\columnwidth]{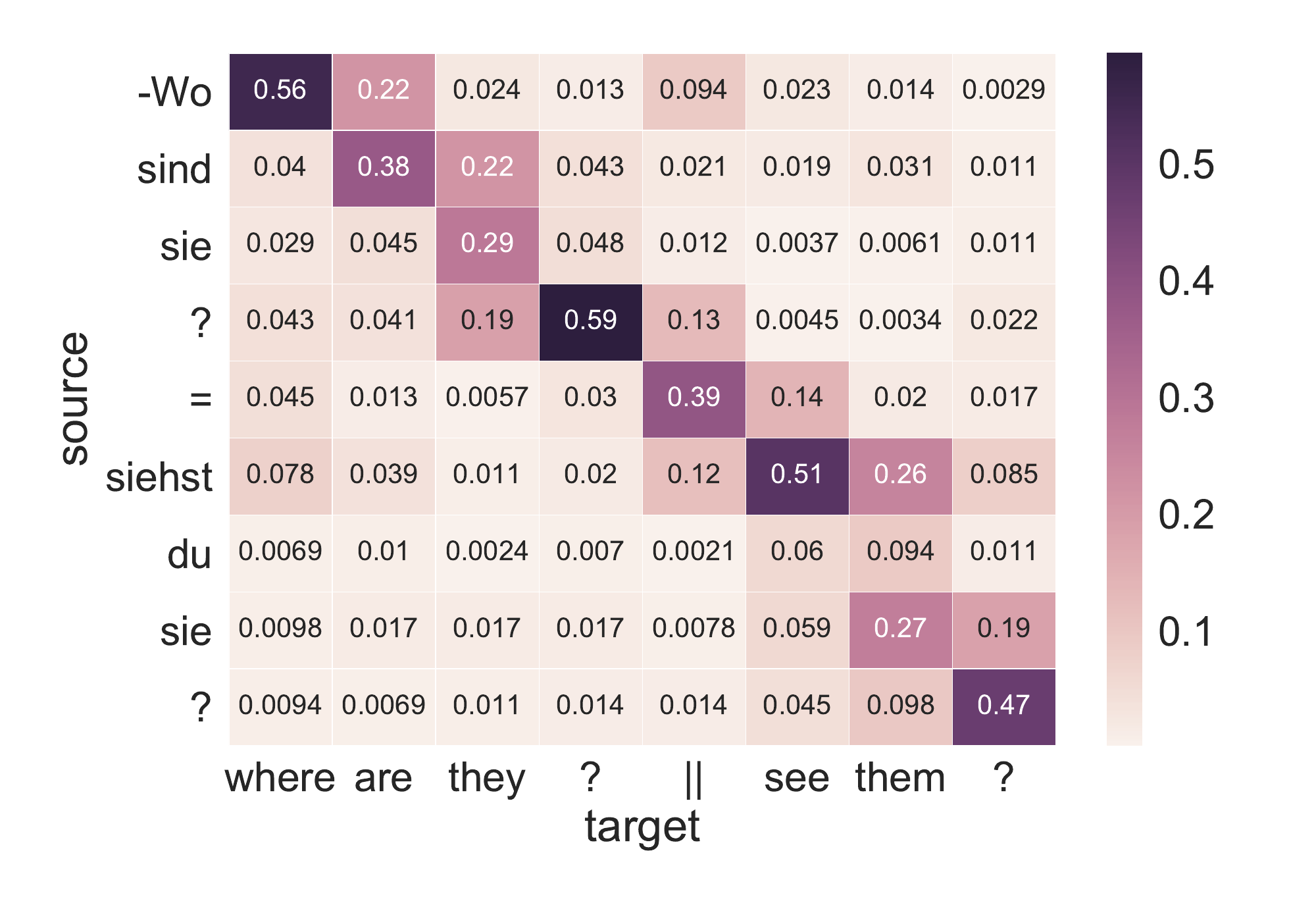}
\caption{Attention patterns with referential pronouns in extended context.}
\label{img:2+2-400}
\end{figure}

We also inspected some translations and their attention distributions in order to study the effect of larger translation units on translation quality. 
One example is the translation in Figure~\ref{img:2+2-400}.

The example illustrates how the model works when deciding translations of ambiguous words like the German pronoun ``sie''. First, when generating ``they'', the model looks at the verb for agreement constraints and the representation around the plural inflection ``sind'' of the German equivalent of ``are'' receives significant attention. Even more interesting is the translation of ``siehst du sie?'', which in isolation is translated to (the intuitively most likely translation) ``do you see {\bf her} ?'' by our baseline model. In the extended model, the translation changes to ``them'', which agrees with the context and is coherent here. Why the auxiliary verb and the subject pronoun are left out is another question but that could be due to the colloquial style of the training data. In any case, the figure shows that ``them'' also looks at ``sind'' in the previous sentence with a weight (0.031) that is significantly larger than for other positions in the previous sentence. This amount seems to contribute to the change to plural, which is, of course, satisfactory in this case. Target language context will certainly also contribute to this effect but even the 2+1 model produces ``them'' in this particular example without the additional target context but the same information from the source.

%
%
%
%

However, sometimes the extended model is worse than the baseline with respect to pronoun translation.  An example is shown below. In this case, the context window is too small and does not cover the important reference ({\em der Sonnenaufgang/the sunrise}), which appears two sentences before the anaphoric pronoun ({\em er/it}). But whether an even larger context model would pick this up correctly is not certain.


\medskip

\noindent
{\footnotesize
\begin{tabular}{|r@{\hskip3pt}p{5.9cm}|}
\hline
context 2:& hast du je den Sonnenaufgang in China gesehen?\\
reference: & ever notice the sunrise in China ?\\
\hline
context 1:& solltest du .\\
reference: & you should .\\
\hline
source: & er ist wunderschön .\\
reference: & it 's beautiful .\\
baseline: & it 's beautiful .\\
extended: & he 's beautiful .\\
\hline
\end{tabular}
}

\medskip

Some translations also become more idiomatic due to the additional context. Empirical evidence is difficult to give but here are three examples that illustrate small changes that make sense:

\medskip


\noindent
{\footnotesize
\begin{tabular}{|r@{\hskip3pt}p{5.9cm}|}
\hline
source: & los , Fenner !\\
reference: & go ahead , Fenner !\\
baseline: & go , Fenner !\\
extended: & come on , Fenner !\\
\hline
\hline
source: & was Sie nicht sagen !\\
reference: & you don 't say !\\
baseline: & what you don 't say !\\
extended: & you don 't say !\\
\hline
\hline
source: & ganz meiner Meinung .\\
reference: & that 's what I say .\\
baseline: & my opinion .\\
extended: & I agree .\\
\hline
\end{tabular}
}

\subsection{Manual Evaluation}

The example of Figure~\ref{img:2+2-400} raises the question whether the extended model is able to reliably and systematically disambiguate pronominal translations. In order to answer this question, we extracted all occurrences of the ambiguous pronoun \textit{sie/Sie} from our test set (1143 occurrences in 1018 sentences, i.e. in every fifth sentence of the test set) and manually evaluated about half of them (565 occurrences in 516 sentences), comparing the output of the baseline system with the one of the 2+2 system. We distinguish four categories on the basis of the reference translation: polite imperative \textit{Sie}, other occurrences of the polite pronoun \textit{Sie}, feminine singular \textit{sie} and plural \textit{sie}. Figure~\ref{tab:manual-eval-sie} lists the results.

\begin{table}[ht]
\centering
{\small
\begin{tabular}{|l|rrr|}
\hline
Word category & Occurrences & Baseline & 2+2  \\
\hline
Polite imperative	& 101 & 98.0\% & 97.0\% \\
Polite other		& 301 & 94.4\% & 95.0\% \\
Feminine singular	& 77  & 85.7\% & 85.7\% \\
Plural				& 86  & 69.8\% & 79.1\% \\
\hline
All					& 565 & 90.1\% & 91.7\% \\
\hline
\end{tabular}
}
\caption{Percentages of correct translations of the pronoun \textit{sie/Sie}.}
\label{tab:manual-eval-sie}
\end{table}

The table shows that polite forms are most frequent in the corpus and also rather easy to translate thanks to capitalisation. In the case of imperatives, they simply are deleted (e.g., \textit{Kommen Sie!} becomes \textit{Come!}), whereas in other contexts they are consistently translated to \textit{you}. The remaining errors are mainly due to entire segments that are left untranslated, or to erroneous lowercasing of sentence-initial positions during preprocessing.

Distinguishing singular from plural readings is harder: a non-polite form \textit{sie} can be translated as \textit{she} or \textit{it} in its singular reading (depending on the grammatical gender of the antecedent), or as \textit{they} or \textit{them} in its plural reading (depending on case). The figures show that the extended model is better at correctly predicting \textit{they} (and \textit{them}), but that correctly predicting \textit{she} or \textit{it} is equally hard with or without context. While the superiority of the 2+2 model cannot be established numerically (none of the reported figures are statistically significant, according to $\chi^2$ tests at $p=0.05$), there are examples that show corrected output:

\medskip

\noindent
{\footnotesize
\begin{tabular}{|r@{\hskip3pt}p{5.9cm}|}
\hline
context: & du bist nur ein Junge und das sind böse Männer .\\
reference: & you 're only a boy , they 're vicious men . \\
\hline
source: & such sie , Max .\\
reference: & get ' em , Max .\\
baseline: & find her , Max .\\
2+2: &find them , Max .\\
\hline
\hline
context: & Sie verstecken sich wie die Ratten im Müll .\\
reference: & they hide out like rats in the garbage .\\
\hline
source: & wenn du sie finden willst , musst du ebenso im Müll wühlen wie sie .\\
reference: & so if you 're gonna get ' em , you 'll have to wallow in that garbage right with them .\\
baseline: & if you want to find her , you 'll have to wallow in the trash like her .\\
2+2: & if you want to find them , you have to dig through the garbage as well as them .\\
\hline
\end{tabular}
}

\medskip

The decision of translating feminine singular pronouns as \textit{sie} or \textit{it} is also improved in some cases by the 2+2 model:

\medskip

\noindent
{\footnotesize
\begin{tabular}{|r@{\hskip3pt}p{5.9cm}|}
\hline
context: & mehr bedeutet dir die Sache nicht ?\\
reference: & is that all my story meant to you ? \\
\hline
source: & was sonst könnte sie mir bedeuten ?\\
reference: & what else could it mean to me ?\\
baseline: & what else could she mean to me ?\\
2+2: & what else could it mean to me ?\\
\hline
\hline
context 2: & kennst du die alte Mine hier ?\\
reference: & know the old mine around here ?\\
\hline
context 1: & - Davon gibt ' s hier viele .\\
reference: & - There 's a lot of them here .\\
\hline
source: & - Sie gehört einem gewissen Sand .\\
reference: & - It 's worked by a man named Sand .\\
baseline: & - She owns a certain sand .\\
2+2: & - It belongs to a certain sand .\\
\hline
\end{tabular}
}

\medskip

However, there is currently not much evidence that these improvements are due to cross-segmental attention. It remains to be investigated if this also holds for the 2+1 model and variants thereof.

\section{Conclusions and Future Work}


In this paper, we present two simple models that use larger context in neural MT, one that adds source language history to the input and one that concatenates subsequent segments in the training data. We discuss the effect on translation and the attention model in particular. We can show that neural MT is indeed capable of translating with wider context and that it also learns to distinguish information coming from different segments or discourse history. We run experiments on German-English subtitle data and we can find various examples in which referential expressions across sentence boundaries can be handled properly. The current study is our first attempt to model discourse-aware neural MT and the outcome is already encouraging. However, evidence so far is rather anecdotal but in the future, we plan to run more systematic experiments with detailed analyses and evaluations. We will look at different windows and other ways of encoding discourse history. We will also study specific discourse phenomena in more depth trying to find out whether NMT learns to handle them in a linguistically plausible way. Finally, this research also intends to provide insights into the development of discourse-aware coverage models for NMT. Indeed, explicit models of coverage have been shown to reduce the amount of overtranslation and undertranslation, whereas our translation models with extended context settings are targeted to make use of overtranslation and undertranslation to some extent. Our experiments will hopefully contribute to a better understanding of the attention and coverage dynamics in discourse-aware NMT.

\section*{Acknowledgments}

We wish to thank the anonymous reviewers for their detailed reviews. We would also like to acknowledge the Finnish IT Center for Science (CSC) for providing computational resources and NVIDIA for their support by means of their GPU grant.

\bibliography{emnlp2017}
\bibliographystyle{emnlp_natbib}

\end{document}